# The Road to On-board Change Detection: A Lightweight Patch-Level Change Detection Network via Exploring the Potential of Pruning and Pooling

Lihui Xue, Zhihao Wang, Xueqian Wang, Member, IEEE, Gang Li, Senior Member, IEEE

This work was supported by National Key R&D Program of China under Grant 2021YFA0715201, in part by National Natural Science Foundation of China under Grants 61925106 and 62101303, and in part by Autonomous Research Project of Department of Electronic Engineering at Tsinghua University.

*Abstract*—Existing satellite remote sensing change detection (CD) methods often crop original large-scale bi-temporal image pairs into small patch pairs and then use pixel-level CD methods to fairly process all the patch pairs. However, due to the sparsity of change in large-scale satellite remote sensing images, existing CD methods suffer from a waste of computational cost and memory resources on lots of unchanged areas, which reduces the efficiency of on-board processing platform with extremely limited computation and memory resources. To address this issue, in this paper we propose a lightweight patch-level CD network (LPCDNet) to rapidly remove lots of unchanged patch pairs in large-scale bi-temporal image pairs. This is helpful to accelerate the subsequent pixel-level CD processing stage and reduce its memory costs. In our LPCDNet, the multi-layer feature compression (MLFC) module is designed to compress and fuse the multi-level feature information at different stages. The output of MLFC module with powerful representations of input patches are fed into the fully-connected decision network to generate the predicted binary label. To compress the model and accelerate the inference speed, a sensitivity-guided channel pruning method is proposed to remove unimportant channels and construct the lightweight backbone network on basis of original ResNet18 network. The performance is better than simply applying fixed pruning ratio to all the stages of the network. In addition, in order to tackle the severe class imbalance problem, a weighted cross-entropy loss is especially utilized in the training process of network. Experiments on two CD datasets demonstrate that our LPCDNet achieves more than 1000 frames per second on an edge computation platform, i.e., NVIDIA Jetson AGX Orin, which is more than 3 times that of the existing methods. In addition, our method reduces more than 60% memory costs of the subsequent pixel-level CD processing stage, significantly lightening the computational burden on edge computation platforms. In addition, most of the changed patch pairs can be correctly selected, ensuring high overall detection accuracy of the entire CD process.

**sensing image, on-board processing, channel pruning, multi-layer feature compression.**

# I. INTRODUCTION

The task of remote sensing change detection (CD) is to identify and segment the changed areas by comparing images of the same region acquired at different times. It is an important technique for the intelligent processing of remote sensing images and serves as a key step for many real-world applications, such as disaster damage assessment [1,2,3], land cover and land use managing [4,5,6], urban expansion surveys [7,8,9], and environmental resources monitoring [10,11]. The remote sensing images used for CD are mainly obtained by imaging equipment on satellites or airplanes. Compared to the way of aerial photography, satellites usually have a larger field of view and more stable imaging conditions.

In most of the existing CD systems, remote sensing images captured by satellites are first downloaded to ground stations and then processed and analyzed to perform CD [12,13]. With the fast development of remote sensing technology, it is easier to obtain all-weather and all-day remote sensing image data. Meanwhile, the resolution and volume of images have rapidly escalated, increasing the pressure on data downlinks [14,15]. The download of data requires a significant amount of bandwidth and time consumption. In this context, the deployment of CD models on satellite edge devices can be an intuitive solution [16,17]. In recent years, driven by the deep learning technologies and large-scale CD datasets with annotations, CNN-based CD methods [18]-[23] have gradually replaced the traditional methods based on algebraic calculation or handcraft-based transformations [24,25] and dominated the field. Although the existing CNN-based methods demonstrate excellent accuracy and robustness, they usually contain billions of multiplication and addition operations. The low efficiency and high computational cost of the network make it difficult to deploy directly on the hardware platforms with limited computing resources or low energy consumption, such as satellites and aircrafts.

In order to achieve on-board processing of high-resolution remote sensing images, the framework of CD for deployment needs to meet the requirements of lightweight, high throughput, and low storage space. In real-world applications, due to the limitation of memory capacity and computing capability, large-size images (>10000×10000) cannot be directly fed into the CD network. Under this circumstance, the existing mainstream framework for large-scale remote sensing images CD works in the following paradigm (see Fig. 1(a)): firstly cropping the large image into many patches with fixed size (e.g., 256×256) by means of sliding windows, then feeding all the patches into the pixel-level CD network, finally merging the CD results of patches to obtain the ultimate large-size change map.

Although the above framework is able to overcome the problem of limited computing resources, large number of image patches still lead to considerable inference time, which is unacceptable for applications. To tackle this issue, most of the existing work focuses on designing lightweight pixel-level CD models to achieve faster inference speed [26,28-31]. Kaiqiang Song et al. proposes an effective network called 3M-CDNet and its lightweight variant 1M-CDNet [26]. The deformable convolution [27] is introduced into the lightweight backbone composed of only a small number of residual modules, improving the ability to extract features of changed areas with irregular shapes. The proposed model exhibits a good trade-off between accuracy and efficiency. In [31], Zhenglai Li et al. adopts a lightweight backbone, MobileNetV2 [32], to extract bi-temporal features, which can reduce the computation costs compared to the commonly-used backbone networks such as ResNet [33], VGG [34], and Unet [35]. In [28], Biyuan Liu et al. incorporates deep separable convolution with different dilation rates into a dual-branch structure and designs the Context Guide Block, which enhances multi-scale features and compresses model complexity at the same time. Considering that low-level features such as lines, points and edges can be general-purpose to all images, TinyCD [29] only uses the features extracted from the first few layers of a pre-trained backbone for subsequent discrimination, which significantly reduces the model size. To reduce the parameters and redundancy of the existing multi-scale feature fusion module, [30] proposes multi-scale decoupled convolution (MSDConv) for feature extraction, which can effectively capture the multi-scale features of changed objects using a compact design with decoupled spatial and channel correlations. An ultralightweight spatial–spectral feature cooperation network (USSFC-Net) is then built based on MSDConv.

Overall, the current methods related to lightweight CD mostly process the images at the pixel level. Although lightweight pixel-level models can compress inference time to a certain extent, large number of patches still occupy a lot of storage space and bring a significant load to data transmission. Actually, in most cases, the changed areas in original large-scale image are usually sparsely distributed, taking up only a small proportion of the entire image. Therefore, most of the image patches do not contain changed areas and it is unnecessary and time-consuming to apply the pixel-level method to all of the patches.

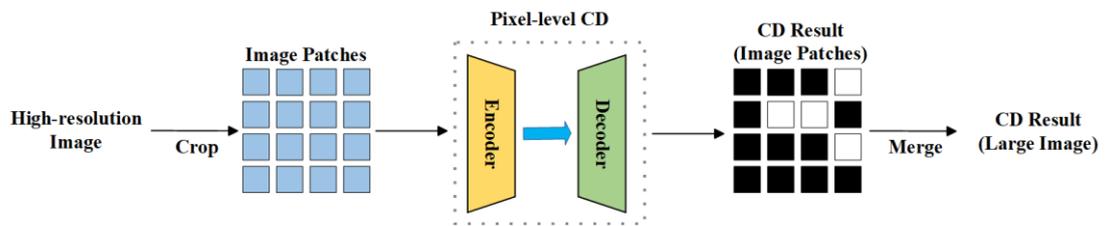

(a) Mainstream Framework (Pure Pixel-level CD)

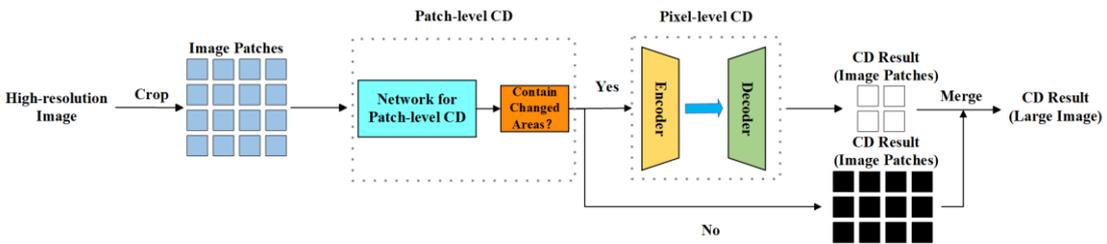

(b) Our Framework (Patch-level CD + Pixel-level CD)

Fig. 1. Comparisons of mainstream framework and ours for CD of large-scale remote sensing images.

To address the above issue, inspired by [36], a good strategy is to select the image patches containing changed areas before applying the pixel-level CD model. Then the pixel-level CD only needs to be conducted on these selected pairs of image patches, which considerably reduces the time cost and the consumption of computing resources. It is crucial for the on-board deployment of CD. Moreover, the volume of data transmitted from on-orbit or on-board platforms (e.g., satellite or UAV) to the ground can be greatly reduced. The process of selecting image patches with changed areas and removing the unchanged patches is precisely the task of patch-level CD. Combining patch-level CD and pixel-level CD, we propose a new framework for large-scale image CD (see Fig. 1(b)), so as to achieve more efficient inference with accurate detection result. In addition, the patch-level CD also demonstrates good robustness to registration error and pseudo changes caused by changes of seasons, lighting conditions, satellite sensors and so on.

To this day, there has been very few studies on patch-level CD. Here we list several existing works with high correlation [36-38]. Based on VGG16 [34], Faize et al. [37] firstly proposes an end-to-end patch-based Siamese neural network for patch-level CD, in which features from various levels are utilized to train the network. However, the lack of feature compression leads to great computational cost, which is unacceptable for real-world deployment. The large dimension of the obtained feature vectors results in a large amount of parameters of the subsequent decision network, making the whole network difficult to converge. With patch-level and pixel-level CD combined, Bao et al. [36] designs an end-to-end network termed PPCNET, which consists of Siamese network for feature extraction, three cascading fully-connected layers and a softmax layer to obtain the final result. It achieves higher accuracy and faster speed than pure pixel-level supervised methods at the time, but the lack of feature fusion at different levels still limits the detection accuracy to a certain extent. In order to solve the problem of misalignment between bi-temporal planetary images and the lack of labeled bi-temporal data in the field of planetary CD, Sudipan et al. [38] proposes an unsupervised patch-level CD method. It exploits a pretrained backbone network with global max-pooling operation to obtain patch-level feature description and adopts self-supervised method to determine the threshold for label prediction. In this method, the global pooling operation for feature compression is only utilized in the last layer of feature extractor, which is prone to lose some detailed spatial information and affect the detection accuracy especially for complex scenes.

On the whole, current patch-level CD methods usually adopt existing backbone networks such as VGG [34] for feature representation, which have relatively large number of basic parameters. Among them, redundant feature channels occupy large proportion of the total parameters and calculations, but have very limited effect on the improvement of detection accuracy. Furthermore, due to the characteristic of patch-level CD task, fine-grained feature extraction at each level is not necessary. In this context, network compression by means of channel pruning can be a good strategy to improve the detection efficiency as well as reducing the processing time without much performance loss. In this paper, to solve the aforementioned problems, we propose a lightweight patch-level CD network based on multi-layer feature compression and sensitivity-guided channel pruning, named as LPCDNet. It has considerably less parameters as well as shorter inference time and achieves a great tradeoff between patch-level detection accuracy and model complexity. The Siamese backbone network of LPCDNet is modified from ResNet18 network [33]. In view of the channel redundancy of original ResNet18, a sensitivity-guided network pruning method is adopted to prune the unimportant channels and construct the lightweight network LW-ResNet18. The sensitivity to pruning operation of each stage in the backbone network is quantified and then utilized to rectify the initial pruning ratio, in order to achieve better pruning performance. For feature extraction, the multi-layer feature compression (MLFC) module is designed to fuse the feature maps acquired at different stages and generate the global feature vector, which is a good representative of the input image patch. Considering the various sizes of changed areas contained in the image patches, the introduction of the multi-scale max-pooling structure helps to improve the detection accuracy. In addition, pooling operations can also reduce the impact of registration errors to some extent, which is meaningful for the practical application of the CD algorithm. The major contributions of our work can be summarized as follows.

- We propose a lightweight network LPCDNet for patch-level CD of optical remote sensing images, which is a pre-selection and efficiency-improved process for the subsequent pixel-level CD. Combining patch-level and pixel-level CD, a new framework for large-scale image CD is designed. Compared to the current mainstream framework, it can considerably reduce the whole processing time as well as maintaining high overall detection accuracy.
- In LPCDNet, the MLFC module is incorporated, which is able to capture and fuse the multi-level feature information at different stages as well as reducing the dimension of feature vectors. To compress the model and accelerate the inference speed, a sensitivity-guided channel pruning method is proposed to remove unimportant channels and construct a lightweight backbone network on basis of ResNet18 network.
- In order to tackle the severe class imbalance problem encountered in patch-level CD task, a weighted cross-entropy loss is adopted to make the training process stable and accelerate the convergence, where the value of weight vector is determined according to the number of changed and unchanged patches.
- Experiments performed on two optical remote sensing datasets for patch-level CD validate the effectiveness and efficiency of our LPCDNet. Experimental results demonstrate that LPCDNet obtains comparable or better performance of patch-level CD than existing competitive methods with fewer parameters, fewer floating-point operations, and higher inference speed. In addition, the framework combining our patch-level CD method and existing pixel-level CD methods shows significant computation and memory resource-efficiencies in comparison with the existing pure pixel-level CD processing methods, which is more friendly to on-board processing platforms on satellites.

The rest of this article is organized as follows. Section II reviews the related works of network pruning and pooling operation. Section III elaborates the details of our proposed LPCDNet approach. Section IV shows in-depth comparisons between the proposed LPCDNet and other state-of-the-art methods, using detailed experimental assessment. Finally, Section V concludes this paper with several remarks and hints at plausible future research.

## II. RELATED WORKS

A. Network Pruning

In order to reduce the high computational cost of convolutional neural networks (CNNs) and realize the on-board deployment, various types of model compression methods have been proposed, including network pruning [39-49], parameter quantization [50-52], low-rank decomposition [53-55], knowledge distillation [56,57], etc. Among them, network pruning is the most popular and extensively studied model compression technique in both academia and industry [47].

Based on the differences in pruning granularity, network pruning methods can be divided into two types: unstructured pruning (or weight pruning) [39-41] and structured pruning (or filter pruning) [42-49]. Unstructured pruning method concentrates on individual weight pruning for the original model according to magnitude or gradient criterions. This strategy usually causes unstructured sparsity patterns. The existing hardware architecture cannot directly accelerate the process of unstructured pruning, and special algorithms need to be designed to support the corresponding sparse operations [39].

Contrarily, structured pruning method prunes part of the network structures (e.g., channels, layers) instead of individual weights. It does not require specialized libraries for sparse computing, which is easy to be implemented and hardware-friendly. Among structured pruning methods, channel pruning is most popular, since it operates at the most fine-grained level while still fitting in conventional deep learning frameworks [42]. To determine the channels to be pruned, some criterion should be utilized to measure the importance of each channel. A commonly-used strategy is to use the norms of filters to evaluate their importance [43,44]. Specifically, the filters with small norm correspond to small values in intermediate output and are identified to be less important than those with large norm values. Considering the limitations of norm-based criterion, the geometric median is selected as the criterion in [45], leading to better pruning performance. In [46], an energy-based filter pruning framework (EFPF) is proposed for channel pruning, where the energy is obtained through the eigenvalues of each weight tensor by singular value decomposition (SVD) technique and can reflect the redundancy of the filters. Compared to the above criterion which is directly calculated from the weights and sensitive to the update of weights, measuring the importance of feature maps can provide a better guideline to determine the important filters [47,48]. In [49], an entropy-based framework is proposed to prune unimportant filters to accelerate and compress CNN models. The channel with small value of entropy is considered to contain less information and can be removed without considerable performance loss. In addition, rank of the feature maps can be also utilized to remove the unimportant channels of network. It is demonstrated in [47] that filters with lower-rank feature maps are less informative and less important to preserve accuracy. Based on this statement, it proposes a channel pruning method, which is mathematically formulated to prune filters with low-rank feature maps.

In order to adapt to the conditions of on-board CD, our paper focuses on the channel pruning to compress the model and accelerate the inference process.

B. Pooling

Pooling operation is widely used in CNNs and mainly used to reduce the spatial size of feature maps. As a downsampling operator, it is able to reduce the computational cost and relieve the overfitting problem. Moreover, pooling operation can also enhance the translation invariance capability [58].

Average pooling [59,60] and maximum pooling [61] are two most common types of pooling operation. They are used to calculate the average and maximum value inside the sliding window of fixed size, respectively. By this mean, the spatial size of the output feature map decreases in comparison to the input. In [62], the above two types are combined through a mixing strategy as well as a gating strategy. Also, a tree-structured self-learning pooling strategy is proposed. With only a slight increase in computational overhead, the designed pooling operations are able to improve the performance and provide a boost in invariance properties relative to conventional pooling operations. [63] replaces the conventional deterministic pooling operations and proposes a stochastic pooling operation, which is a novel type of regularization for CNNs. The stochastic pooling calculates the activation from a multinomial distribution formed by the values in the pooling window. It prevents the training process of large models from over-fitting and shows superior performance. Lin et al. designs a new type of pooling termed global average pooling and utilizes it to replace the traditional fully-connected layers in CNNs [64]. The global average pooling is more native to the convolution structure and contains no parameter to optimize.

In addition to the methods mentioned above, some other advanced pooling methods have been investigated, such as soft pooling [65], local importance-based pooling [66], and strip pooling [67]. In this paper, we only use the basic pooling operation to compress the obtained feature maps with high dimension, so as to achieve fast inference and low algorithm complexity.

## III. PROPOSED LPCDNET METHOD

In this section, the network architecture of the proposed LPCDNet for patch-level CD is firstly introduced. With a pair of bi-temporal image patches as input, the goal of our LPCDNet is to detect whether it contains changed areas and output a binary label. Different from the existing networks for pixel-level CD which assign binary label to each pixel, the output label of LPCDNet is patch-level. Next, we elaborate on the proposed MLFC module consisting of max-pooling layers at multiple stages of the backbone network, which is utilized to compress and fuse the feature information with different levels. Afterwards, the sensitivity-guided network pruning method is illustrated in detail. It is adopted to build our lightweight backbone network LW-

ResNet18 (modified from ResNet18 [33]).

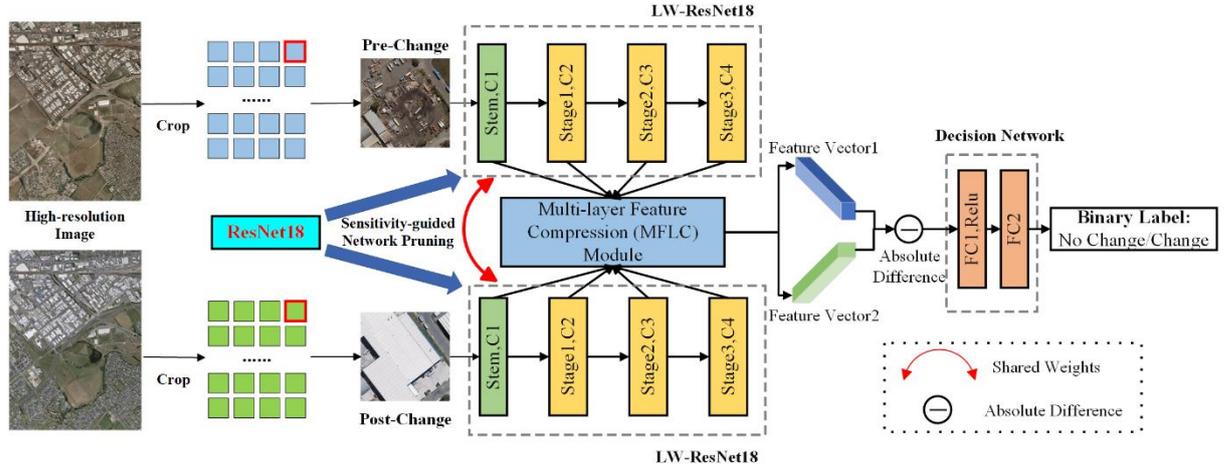

**Fig. 2. Network architecture of the proposed LPCDNet.**

A. Network Architecture

According to the definition stated in section I, the patch-level CD can essentially be regarded as an image classification task with two classes. Nevertheless, owing to the characteristic of remote sensing images such as the existence of registration error and various sizes of changed areas, the patch-level CD is much more complicated than conventional image classification tasks. Under this circumstance, we still utilize the overall architecture of network similar to that used for traditional image classification task, including an encoder for feature extraction and a prediction head for classification. Simultaneously, to overcome the challenges encountered in patch-level CD, some modifications are adopted to achieve good performance. The architecture of the proposed LPCDNet is shown in Fig. 2. It consists of three parts: a Siamese encoder based on lightweight network LW-ResNet18, the MLFC module to refine and fuse the multi-scale feature maps, and a two-layer decision network for final decision.

The changed areas in CD task usually have various scales and locations. In this context, method focusing on detection of multi-scale changed areas can achieve great performance. For this purpose, we propose the MLFC module to generate a global feature vector of the bi-temporal patch pair. Feature maps from different levels of the backbone network are fed into the module, in which point-wise convolution and max-pooling operation are used to compress the feature maps into 1-D vector. The obtained 1-D vectors for each stage are concatenated to form the output feature vector, which contains multi-scale feature information and serves as a good representative of the input. Subsequently, the absolute difference between the feature vectors of bi-temporal input images is calculated. The result contains various levels of information related to changed areas, which is beneficial to patch-level CD. Eventually, the difference vector is fed into the decision network to generate the binary patch-level classification result.

As for the backbone network, ResNet18 network is selected in this paper as the base model. Compared with other commonly-used networks such as VGG [34] and UNet [35], ResNet [33] adopts the residual connection, which helps to avoid gradient vanishing problem and improve the network performance. Moreover, in terms of parameters (Params.) and multiply-accumulate operations per second (MACs), the model volume of ResNet18 is smaller, leading to better model efficiency. To further simplify the original ResNet18 network and accelerate the inference process, a sensitivity-guided network pruning method is proposed to use.

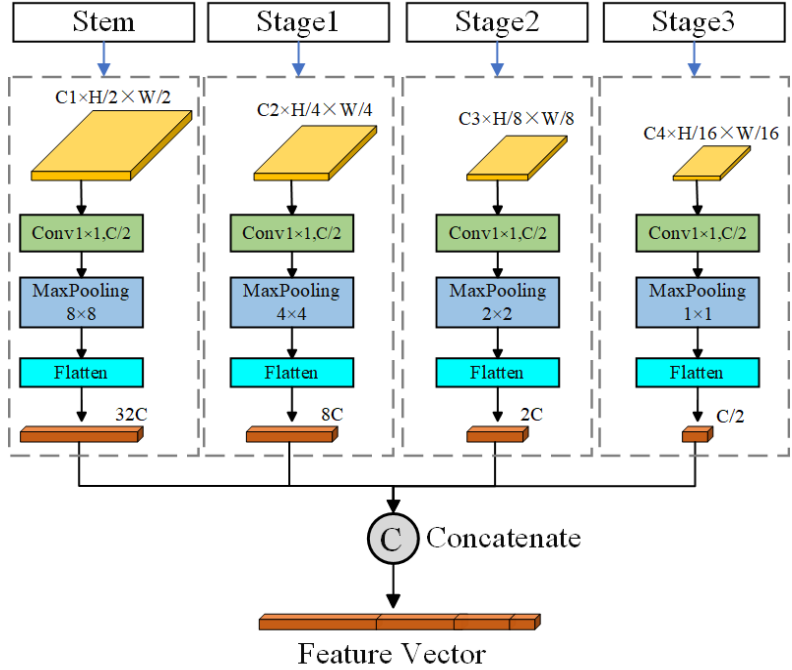

**Fig. 3. Structure of the proposed MLFC module.**

B. Multi-layer Feature Compression

In traditional image classification network, the output feature map of last layer in encoder part is usually flattened into a 1-D vector using global pooling and then fed into the subsequent prediction head for classification. For patch-level CD task, the changed areas contained in the image patches have varied sizes and the difference of sizes can be significant. Therefore, it is not appropriate to only utilize the features of last stage for final prediction, which will limit the detection accuracy. To tackle this issue, the strategy of fusing the features from multiple stages can be an effective way, which utilizes various levels of feature information at different stages and fuses them to generate a global feature vector as a great representative of input.

Meanwhile, to minimize model complexity as much as possible while ensuring sufficient detection accuracy, the max-pooling layer is adopted before fusion at each stage. It can not only retain important features but also reduce computation and avoid overfitting, thus improving the model's generalization ability. Furthermore, it also maximizes the translation invariance of the feature map, which is able to reduce the impact of registration errors on detection results to some extent. In view of the granularity difference of feature information, we avoid using global pooling at each stage, which leads to considerable loss of fine-grained features in low-level stage. Instead, fixed size of pooling window is assigned to all of the pooling layers. With this approach, the spatial size of output in low-level stage can be large enough to preserve sufficient detailed information.

On basis of the above strategies, we build the MLFC module and feed the feature maps from multiple layers extracted by backbone network to it. The structure of it is demonstrated in Fig.3. In the proposed module, we apply 1×1 convolution and max-pooling operation at all of the four stages (including the stem block), compressing the feature maps in channel and spatial dimension respectively. Subsequently, the compressed feature maps are flattened into 1-D feature vectors, which are then concatenated to obtain the global feature vector. After fusion, the obtained global feature vector contains multiple levels of feature information of ground objects in the image patch, which will be useful in the discrimination of changed areas. The corresponding calculation process can be formulated as follows.

$$V_i = Flatten\left(F_{maxpool,i}\left(f^{1\times1,C}(U_i)\right)\right), i = 1,2,3,4$$
$$V_{out} = Concat([V_1, V_2, V_3, V_4]) \quad (1)$$

where $U_i$ is the output feature maps of stage i (i=1,2,3,4) and $V_i$ denotes the obtained 1-D feature vector at corresponding stage. $F_{maxpool,i}$ represents max-pooling operation and $f^{1\times1,C}$ denotes the standard convolution operation with $1 \times 1$ kernel size as well as C output channels. The value of C is set to half of the minimum channel number of all stages. $V_{out}$ is the global feature vector output by MLFC module and $Concat[\cdot,\cdot]$ here denotes concatenating two 1-D feature vectors. With input size of $128 \times 128$, the spatial sizes of the output feature maps at four stages are $64 \times 64$, $32 \times 32$, $16 \times 16$ and $8 \times 8$, respectively. In max-pooling layer, the size of sliding window is set to $8 \times 8$ for each stage. Correspondingly, the output sizes of max-pooling layers for four stages are determined as $8 \times 8$, $4 \times 4$, $2 \times 2$ and $1 \times 1$, respectively. Ultimately, for simplicity, the multiple feature vectors V1, V2, V3, V4 are directly concatenated to form the global feature vector $V_{out}$, which is then fed into the following decision network.

By incorporating the MLFC module, multi-scale features of the ground objects from different stages are refined and fused, generating a 1-D feature vector with great representation of input image patch. Aimed at detecting the changed areas of various sizes, the use of multi-layer feature compression is able to obtain more accurate detection results in complex scene.

## C. Sensitivity-guided Network Pruning

In this subsection, we introduce the sensitivity-guided network pruning method, which is adopted to build the lightweight backbone network LW-ResNet18.

Typical lightweight networks used in the field of CD include MobileNet [32,68,69], GhostNet [70], ShuffleNet [71], etc. In the aforementioned lightweight networks, group convolutions and depth-wise separable convolutions are commonly adopted as alternatives to standard spatial convolutions, so as to reduce the parameters and floating-point operations per second (FLOPs) of deep networks. In theory, depth-wise convolution requires less computation, however its arithmetic intensity (ratio of FLOPs to memory accesses) is too low to efficiently utilize hardware [72]. Therefore, although the model complexity (measured by parameters and FLOPs) is considerably reduced after applying the lightweight convolutions such as depth-wise separable convolutions, the inference process cannot be accelerated significantly as expected. To this end, we abandon the use of typical lightweight networks and build a new lightweight backbone network modified from ResNet18 by means of network pruning. Compared with commonly used networks such as VGG and UNet, ResNet adopts residual connections to avoid gradient vanishing problems and has better model efficiency. Therefore, ResNet18, which is the smallest model in ResNet family, is selected as the base model.

*Channel Redundancy:*

Fig. 4 demonstrates partial channels of the feature maps of a certain image patch extracted from the first stage of original ResNet18, before and after simple channel pruning operation (compression ratio is set to 1/8). It visualizes the redundancy of feature maps generated by ResNet18 in channel dimension. Fig. 4 shows that for a single ground object (e.g., building and road) in the image, the visualization results of feature maps from different channels can be similar, leading to obvious redundancy among different channels. In addition, as shown in (d), the feature information related to ground objects contained in some channels is very limited (e.g., part of the boundary), but the extraction of them requires a significant amount of parameters and computation. Therefore, it is feasible to compress the patch-level CD model by means of channel pruning. In Fig. 4, when the number of feature channels is reduced from 64 to 8, most of the detailed information of objects in the input image patch can still be preserved well in the output feature maps. This ensures that the channel pruning method can be applied to the original backbone network in patch-level CD, without marked performance loss.

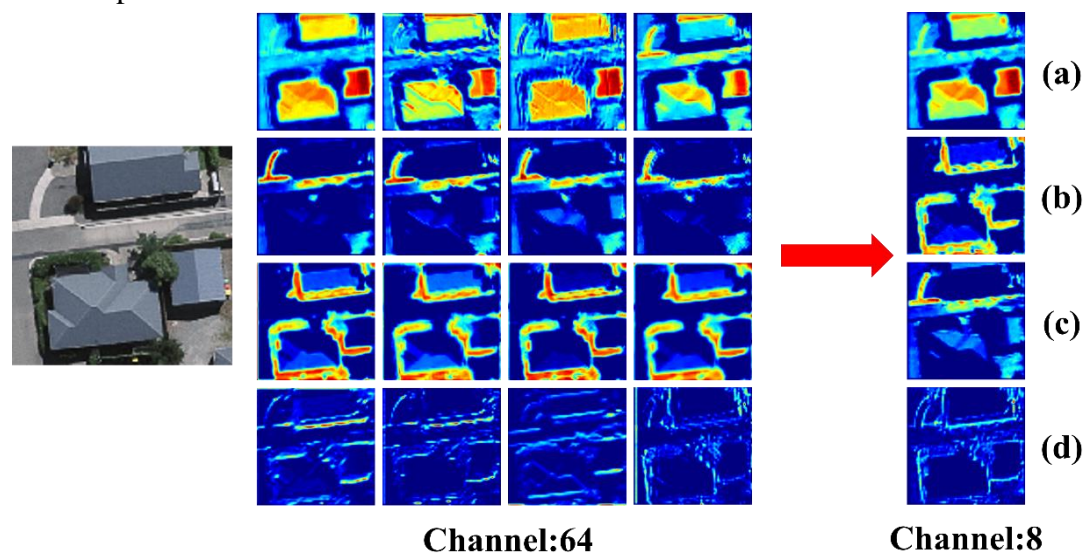

**Fig. 4. Visualization of the redundancy of feature maps generated by ResNet18 in channel dimension.**

*Network Pruning Method:* In view of the channel redundancy of original ResNet18, a sensitivity-guided network pruning method is adopted to prune the unimportant channels and compress the feature extraction network, which is hardware-friendly and easy to implement. Without destroying the model structure, it can be accelerated by Graphics Processing Unit (GPU) or other hardware to obtain faster inference speed.

For network pruning, the number of pruned channels or the value of pruning ratio should be determined manually in most cases. A simple and widely-used way is to assign a fixed pruning ratio to each layer of the base model. However, different layers in backbone network contain different levels of feature information and exhibit different sensitivity to the pruning operation. As a consequence, the strategy of adopting identical pruning ratio in each layer is abandoned. Instead, we propose to quantify the sensitivity of layer to pruning operation and utilize it to rectify the initial pruning ratio, which forms a sensitivity-guided network pruning method.

The main steps of our sensitivity-guided network pruning method are shown in Algorithm. 1. Wherein the defined sensitivity function is:

$$S_i = F(l_i) = \frac{0.5}{1 + \exp\left(\alpha \cdot \frac{l_{max} + l_{min} - 2l_i}{l_{max} - l_{min}}\right)} \quad (2)$$

where $\alpha$ is a positive constant used to control the slope of function. $l_{min}$ and $l_{max}$ refer to the minimum and maximum performance loss among the four pruned networks corresponding to four stages. After obtaining the sensitivity value, the new channel number $C_i'$ can be calculated by:

$$C_i' = C_i \cdot (1 - \lambda_i) = C_i \cdot (1 - s_i \cdot \lambda) \quad (3)$$

**Algorithm 1** *Sensitivity-guided Pruning Method*

Input: original ResNet18 network, initial pruning ratio $\lambda$, sensitivity function $F$
Output: rectified pruning ratio $\lambda_i (i = 1,2,3,4)$, pruned network LW-ResNet18
(1) Train the original patch-level CD network with absence of MLFC module and store values of the trained filters.
(2) Use the initial pruning ratio to prune the filters in all of the four stages in sequence. $c_i\lambda$ filters with smallest values of L1-norm are pruned.
(3) Using the pruned filters in (2) for initialization, retrain the corresponding pruned network. Calculate the performance loss $l_i (i = 1,2,3,4)$ compared to original model.
(4) Compute sensitivity value $s_i (i = 1,2,3,4)$ of each stage in ResNet18 using the defined sensitivity function $F$.
(5) Update the initial pruning ratio $\lambda$ and obtain the rectified ratio $\lambda_i$ for each stage.
(6) Calculate the channel number of each stage after pruning on basis of $\lambda_i$ and build the pruned LW-ResNet18 network.
(7) Build the pruned LPCDNet consisting of the LW-ResNet18, the MLFC module and decision network. Train from scratch and measure the detection performance.

The whole process is modified from the max response selection strategy with sensitivity function proposed in [73]. The difference is that we calculate the accuracy loss and sensitivity at each stage rather than each layer, in order to reduce the complexity and runtime of network pruning algorithm. According to [43], for deep networks such as VGG or ResNet, layers in the same stage (with the same feature map size) have a similar sensitivity to pruning. To achieve more convenient network pruning, we use the same pruning ratio for all layers in the same stage.

In the experiment, we observe that if the pretrained parameters after pruning are directly used for forward inference, the detection performance will deteriorate sharply and the corresponding sensitivity value can be unreliable. Therefore, instead of directly using the pruned weights to calculate the accuracy loss, we retrain the model until convergence, with the aforementioned pruned weights for initialization. In the process, L1-norm is used to reflect the relative importance of each filter, similar to [43]. In comparison with other criterions such as rank, entropy and energy, it is more convenient and efficient to prune using norm value, without considerable reduction of the detection accuracy.

It has been proved in [74] that if pruning channels in a specific conv-layer leads to more accuracy loss than pruning other layers, namely this layer is more sensitive, reserving all channels of the sensitive layers achieves better compression performance. Thus, for layers sensitive to pruning, we decrease the initial pruning ratio to reduce the performance loss. While for layers not sensitive to pruning, we adopt a pruning ratio close to the initial value.

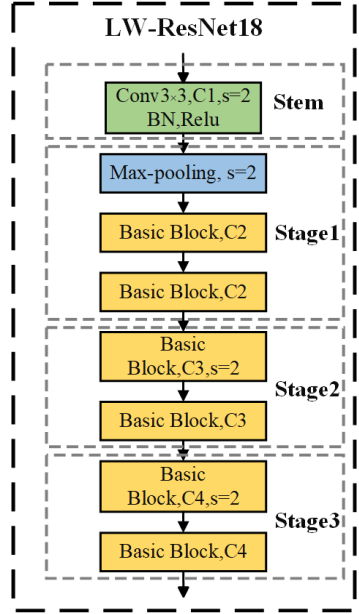

**Fig. 5. Network architecture of the proposed backbone network LW-ResNet18.**

*LW-ResNet18:* With the channel pruning strategy stated above, the lightweight backbone network LW-ResNet18 can be constructed on basis of ResNet18, as shown in Fig. 5. The overall framework as well as the structure of each stage is completely preserved. Concretely, similar to the ResNet18, the LW-ResNet18 network contains a stem block composed of a single convolutional layer, and three cascaded stages each of which contains two basic blocks proposed in ResNet18. The last stage of the original ResNet18 model is removed, in order to avoid redundancy of high-level feature information in deep layers. The total downsampling rate of the network is set to 16 and the number of channels at each stage is calculated using Formula (3). Combined with the proposed MLFC module, the designed LW-ResNet18 compresses the model to a great extent without serious degradation of the detection accuracy.

### D. Weighted Cross-entropy Loss

As stated above, the output of our network for patch-level CD is the probability of containing changed areas. Therefore, the commonly-used cross-entropy loss in the field of classification task can be adopted to train our network. Considering the sparsity of the changed areas in most cases, the number of changed patches is usually much smaller than that of the unchanged patches, leading to severe class imbalance problem which makes the training process unstable and slows down the convergence of network. Under this circumstance, the simple cross-entropy loss without category weights is not appropriate for our network. Instead, the weight vector is additionally introduced into the cross-entropy loss function, which assigns higher value of weight to changed patches. This is helpful to alleviate the problem of class imbalance. The formulation of the proposed weighted cross-entropy (WCE) loss is:

$$L_{wce} = -\frac{1}{N}\sum_{k=1}^{N}\sum_{i=0}^{1} w_i \cdot y_{i,k} \cdot \log\left(\frac{e^{\hat{y}_{i,k}}}{\sum_{l=0}^{1} e^{\hat{y}_{l,k}}}\right) \qquad (4)$$

where N is the total number of image patches in the training dataset, $i \in \{0, 1\}$ denotes the class of the image patch (i.e., $i = 0$ represents an unchanged patch, $i = 1$ represents a changed patch), $\hat{y}_{i,k}$ represents the predicted probability of class $i$ of the $k$-th patch, $y_{i,k}$ represents the ground truth label of the $k$-th patch, and $w$ is the weight vector of the WCE loss which satisfies $w_0 + w_1 = 1$. In our implementation, the value of weight vector $w$ is determined according to the number of changed and unchanged patch pairs in the training data. Denoting the number of changed patches as $N_{change}$, then the corresponding weight vector is: $w = (\frac{N_{change}}{N}, 1 - \frac{N_{change}}{N})$.

## IV. EXPERIMENTAL RESULTS AND ANALYSIS

### A. Experimental Setup

*1) Dataset:* To test the performance of the proposed LPCDNet, we conduct experiments on two datasets for patch-level CD of optical remote sensing image, i.e., WHU-128 and GZLandslide-128.

**WHU-128:** Modified from the large-scale aerial images for building CD proposed in [75], WHU-128 dataset is specially built by us to test the patch-level CD algorithms. The original dataset contains two pairs of bi-temporal high-resolution(0.075m) aerial images of adjacent areas in Christchurch, New Zealand. The size is 21243 × 15354 for one pair and 11265 × 15354 for another. In addition, the changed areas are sparsely distributed, which satisfies the conditions for the use of patch-level CD. Based on the above large images, WHU-CD dataset is built to test the pixel-level CD methods. Specifically, images from two pairs are concatenated into a 32508 × 15354 large image and then cropped into many small image patches with size 256 × 256 with no overlap. The obtained image patches are randomly split into three subset for training/ validation/test, respectively.

For our WHU-128 dataset, the size of image patch is set to 128 × 128 and the cropping operation is conducted with 50% overlap. Different from the generation of WHU-CD, the large image with size 21243 × 15354 is used to obtain the training and validation dataset, while the patches of test dataset are derived from the other large image. WHU-128 dataset contains 121173 pairs of remote sensing image patches in total, with 63287/15822/42064 pairs for training/validation/test respectively. It contains building changes of various shapes and sizes, so that it can be adopted to validate the effectiveness of the proposed MLFC module.

**GZLandslide-128:** GZLandslide-128 dataset is an optical remote sensing dataset for patch-level CD of landslide disaster occurred in Bijie, Guizhou on August 28, 2017. The original pair of large-scale images includes one taken by Gaofen-1 satellite before the disaster and another taken by Gaofen-2 satellite after the disaster, with size of $51729 \times 58505$. The resolutions of the bi-temporal images are 1m and 2m respectively. Considering that the landslide areas only take up a very small part of the entire image, we use part of the original large image to build the dataset, in order to alleviate the problem of imbalance between the samples of negative and positive class.

In detail, we cut out a sub-image of size $4096 \times 4096$ containing all of the landslide areas to verify the performance of patch-level CD methods. Subsequently, the selected sub-image is cropped into 128×128 patches with 50% overlap. To increase the number of patches containing changed areas, data augmentation is adopted to reduce the impact caused by category imbalance. Ultimately, the generated GZLandslide-128 dataset totally contains 4713 pairs of image patches of size $128 \times 128$, in which training/validation/test dataset includes 3293/468/952 pairs respectively.

Note that the above two patch-level CD datasets both provide pixel-level change labels. But in patch-level CD task, the pixel-level label is only used to generate a binary label indicating whether the patch includes changes or not. The training process of patch-level CD method only utilizes the obtained binary label, without the pixel-level information of changed areas.

*2) Evaluation Metrics:* Similar to classical pixel-level CD, the performance of patch-level CD can be described by confusion matrix. Therefore, indicators used in pixel-level CD such as precision, recall, F-score can also be applied to measure the performance of patch-level CD method. In this paper, evaluation metrics mainly include recall, F-score, Matthews correlation coefficient(MCC) and PatchAcc proposed by us. For patch-level CD, the goal is to remove the unchanged image patches as much as possible, while ensuring a high detection rate of changed patches in the meantime. This can be directly described by two indicators: the recall of positive class(with changed areas) and negative class(without changed areas).

Taking both precision and recall into account, F-score is a synthetical indicator, where $\beta$ in the calculation formula is used to balance between the two components. If $\beta = 1$, the F-score is named as F1-score, which is widely used in pixel-level CD as a core metric. In this paper, the value of $\beta$ in F-score is roughly determined according to the number of samples of the two classes in the dataset. As for patch-level CD, it is more important to correctly detect most of the changed patches, in order to maintain the overall accuracy of the entire CD process. Thus, higher weight is assigned to the recall of positive class. Additionally, based on the recall of two classes, we propose a comprehensive indicator, termed as PatchAcc, with the value ranging from 0 to 1. The calculation of PatchAcc is similar to that of F-score. Simultaneously considering the detection rate of changed image patches and the removal rate of unchanged image patches, it is able to comprehensively describe the performance of patch-level detection. Compared to F-score, it reduces sensitivity to the removal rate of unchanged patches and assigns more importance to the detection rate of changed patches, which conforms to the requirements of patch-level CD. For MCC, it is an indicator used to measure the performance of binary classification, taking TP, TN, FP and FN into account. The value ranges from -1 to 1. The reason for selecting it is that MCC is suitable for datasets with imbalanced sample numbers of different categories, which is common in patch-level CD task. The above indicators can be calculated as follows:

$$Precision_{pos} = \frac{TP}{TP + FP}$$
$$Recall_{pos} = \frac{TP}{TP + FN}, Recall_{neg} = \frac{TN}{TN + FP}$$
$$F_\beta = \frac{\beta + 1}{\beta \cdot Recall_{pos}^{-1} + Precision_{pos}^{-1}} \quad (5)$$
$$PatchAcc = \frac{\beta + 1}{\beta \cdot Recall_{pos}^{-1} + Recall_{neg}^{-1}}$$
$$MCC = \frac{TP \cdot TN - FP \cdot FN}{\sqrt{(TP + FP)(TP + FN)(TN + FP)(TN + FN)}}$$

where TP, TN, FP and FN denote the number of pixels of true positive, true negative, false positive and false negative class, respectively. $Recall_{pos}$ and $Recall_{neg}$ represent the recall of positive class and negative class. $Precision_{pos}$ is the precision of positive class. Note that larger values of the above metrics indicate better performance for patch-level CD.

*3) Implementation Details:* Our model is implemented on Pytorch framework and trained using a single NVIDIA RTX TITAN GPU with 24 GB memory. In the experiment, the initial compression ratio for network pruning is empirically set to 0.125. For network training, we use the stochastic gradient descent (SGD) with momentum to update the parameters and optimize the model. The momentum is set to 0.99 without weight decay in the experiments. The training period is set to 90 epochs for WHU-128 and 60 epochs for GZLandslide-128, respectively. The learning rate is initially set to 0.0001/0.001 for WHU-128/GZLandslide-128 and decreases by a factor of 10 after every third of the total training epochs. For weighted cross-entropy loss in the training process, the weight vector for WHU-128 and GZLandslide-128 is (0.15, 0.85) and (0.18, 0.82) respectively. After each epoch of the training process, the model is validated on the validation set, with PatchAcc calculated to reflect the generalization performance of the current model. In fact, F-score can also be used to describe the generalization performance. But since the two recall rates included in PatchAcc are directly related to the detection and removal rates which patch-level CD task focuses on, PatchAcc is chosen here to measure the generalization ability. The model with highest PatchAcc on the validation set is selected as the optimal model, which is then evaluated on the test set to obtain the final values of metrics.

For test, the proposed patch-level CD method is deployed not only on a Windows server equipped with an Intel Xeon Platinum 8280 Central Processing Unit (CPU) and NVIDIA TITAN RTX GPU, but also on NVIDIA Jetson AGX Orin, which is a small embedded AI computer with deep learning accelerators. Integrating NVIDIA Ampere architecture GPU and Arm Cortex-A78AE CPU, as well as the new generation of deep learning and visual accelerators, Jetson AGX Orin can perform 200 trillion operations per second (TOPS), which is comparable to servers with built-in GPUs. Its small size, low power consumption, and high integration make it suitable for on-board processing on platforms such as satellites and airplanes. Earlier, the previous generation of AGX Orin product has been used to verify the performance of real-time on-board detection algorithms [76]. Therefore, we utilize it to simulate the on-board deployment environment and evaluate the efficiency of our proposed method.

It is worth noting that the training process of network for patch-level CD usually exhibits stronger randomness than pixel-level CD, due to the limited supervisory information used in training process. To reduce randomness, we conduct all of the experiments for three times and use the mean value of indicators as the final results.

### B. Performance and Efficiency Comparison

To evaluate the proposed LPCDNet, comparative experiments should be conducted. Specifically, existing patch-level CD methods in [36] and [38] are selected for comparison. Since the existing work related to patch-level CD is relatively limited, several networks for remote sensing image scene classification are also adopted and modified for comparison [77],[78]. Additionally, inspired by [38], two conventional CD methods, i.e., RCVA [79] and SVM [80], are also modified and selected as comparison methods. All of the comparison methods mentioned above are as follows:

1) RCVA [79]: RCVA is originally proposed for pixel-level prediction. To adjust to the task of patch-level CD, we calculate the magnitude of the vector corresponding to each pixel and obtain the magnitude image. Then it is fed into a global max-pooling layer and compared with the selected threshold for ultimate classification. The validation set is used to search for the optimal threshold.

2) CNN-SVM [80]: CNN-SVM integrates the CNNs for feature extraction and the SVM for classification. The Siamese network consisting of two VGG16 networks with shared weights is utilized to obtain the feature maps containing characteristics of the ground objects. Subsequently, feature compression is conducted on the feature maps and the generated feature vectors are then fed into the cascading SVM classifier. The SVM uses Gaussian radial basis function (RBF) as the kernel function and the value of C and $\gamma$ is determined through grid search.

3) PUPCD(VGG16/ResNet18) [38]: Patch-Level Unsupervised Planetary CD (PUPCD) is originally an unsupervised method for patch-level CD and the threshold inside is acquired by means of self-supervised strategies. Considering that the LPCDNet to be compared is a supervised method and the binary labels for network training are available, the original self-supervised method is abandoned. Instead, the value of threshold is determined based on the training and validation dataset, which is similar to the generation of threshold in RCVA. In comparative experiments, VGG16(the first six convolutional layers) and ResNet18 networks are successively used for deep feature extraction.

4) SE-MDPNet [77]: SE-MDPNet uses MobileNetV2 as the base network and introduces the dilated convolution as well as channel attention to extract discriminative features. Furthermore, a multi-dilation pooling module is incorporated after the last stage of the backbone network to extract multi-scale features and improve the detection performance. In order to adapt to the bi-temporal input in patch-level CD task, the original single-input feature extraction network is modified into Siamese network with shared weights, and the difference map between the dual-branch output feature maps is used for subsequent discrimination. To compress the model complexity and avoid the overfitting problem, we set the channel compression ratio of the MobileNetV2 to 0.25, instead of the default value 1.0.

5) ESPA-MSDWNet [78]: Same as SE-MDPNet, ESPA-MSDWNet also adopts MobileNetV2 as the backbone, with channel compression ratio set to 0.25. In addition, the multi-scale depth-wise convolution (MSDW Conv) and efficient spatial pyramid attention (ESPA) module are proposed to improve the capability of feature representation. By use of the MSDW Conv, it is able to represent features in remote sensing images at a more fine-grained level and expand the receptive fields. Moreover, ESPA module can help to extract dependencies between channels. The modification strategy of changing single network into Siamese network is identical to that in SE-MDPNet.

TABLE I
COMPARISON RESULTS ON WHU-128 AND GZLANDSLIDE-128. THE HIGHEST SCORE IS MARKED IN BOLD. ALL THE SCORES ARE DESCRIBED IN PERCENTAGE(%).

| Method | WHU-128 Rec.(neg)/Rec.(pos)/MCC/PatchAcc/F-score | GZLandslide-128 Rec.(neg)/Rec.(pos)/MCC/PatchAcc/F-score |
|---|---|---|
| RCVA [79] | 33.75 /89.86 /17.37 /72.61 /56.01 | 68.40 /93.44 /49.16 /87.61 /76.00 |
| CNN-SVM [80] | 47.81 /**95.30** /29.62 /83.46 /64.36 | 56.05 /84.70 /32.13 /77.50 /64.76 |
| PUPCD(VGG16) [38] | 38.38 /90.75 /20.78 /75.95 /58.09 | 18.60 /93.44 /12.85 /53.96 /58.04 |
| PUPCD(ResNet18) [38] | 21.28 /95.07 /14.11 /63.57 /55.01 | 13.91 /91.80 /6.74 /45.50 /55.88 |
| SE-MDPNet [77] | 74.17 /90.47 /46.16 /87.71 /73.70 | **98.35** /88.52 /**88.45** /90.16 /89.26 |
| ESPA-MSDWNet [78] | 73.72 /91.06 /46.30 /88.05 /73.90 | 86.13 /92.35 /68.51 /91.03 /84.50 |
| LPCDNet(MobileNetV2) | 69.73 /90.64 /42.18 /86.89 /71.42 | 86.30 /85.25 /63.21 /85.42 /79.08 |
| LPCDNet(ShuffleNetV2) | 70.94 /90.02 /42.90 /86.65 /71.62 | 88.08 /91.26 /70.32 /90.66 /84.88 |
| **LPCDNet** | **77.86** /90.47 /**50.11** /**88.41** /**75.84** | 94.50 /**97.45** /86.03 /**96.88** /**93.94** |

6) LPCDNet(MobileNetV2/ShuffleNetV2): In this method, the proposed LW-ResNet18 network is replaced by MobileNetV2/ShuffleNetV2 and the MLFC module is removed. By comparing with our proposed LPCDNet, the effectiveness of the ultra-lightweight backbone network LW-ResNet18 and the multi-layer feature compression strategy can be validated.

Table I demonstrates the comparison results on WHU-128 and GZLandslide-128. In the table, the β value in F-score for WHU-128 and GZLandslide-128 is 6 and 4.5 respectively, which is determined according to the number of positive and negative samples in the training set. The quantitative results indicate that our LPCDNet outperforms the other state-of-the-art comparison methods in terms of MCC, PatchAcc and F-score with a significant margin and reaches the best performance. Concretely, our LPCDNet is able to correctly detect 90.47% and 97.45% of the changed image patches on WHU-128 and GZLandslide-128 respectively. Meanwhile, 77.86% and 94.50% of the unchanged patches can be correctly identified and removed, which can commendably meet the requirements of patch-level CD task. Additionally, we can observe that the indicators of threshold-based methods are considerably lower than the end-to-end methods without threshold for final discrimination. The robustness as well as generalization ability of threshold-based methods is also inferior, especially when there is a significant difference between the distribution of training data and test data.

For comparison of the model efficiency, all of the methods stated above are tested on a Windows server and the AGX Orin. Table II reports the value of Params., MACs, and frames per second (FPS) of these methods. The FPS metric here represents the number of patch pairs that can be processed per second. The values of Params. and MACs obtained on AGX Orin are identical to those obtained on Windows server, while FPS on the two platforms are different, as shown in Table II.

TABLE II
EFFICIENCY COMPARISONS AMONG DIFFERENT METHODS. MACS IS CALCULATED WITH INPUT SIZE OF 128 × 128 × 3 AND FPS IS CALCULATED WITH BATCH SIZE EQUAL TO 64. 1G IS EUQAL TO $10^9$ AND 1M IS EQUAL TO $10^6$.

| Method | Params. | MACs | FPS* | FPS** |
|---|---|---|---|---|
| RCVA [79] | - | **3.15M** | 10 | 9 |
| CNN-SVM [80] | 14.72M | 10.05G | 599 | 94 |
| PUPCD(VGG16) [38] | 1.15M | 9.82M | 46 | 38 |
| PUPCD(ResNet18) [38] | 2.78M | 1.84G | 407 | 202 |
| SE-MDPNet [77] | 367.12K | 167.71M | 1341 | 572 |
| ESPA-MSDWNet [78] | 1.06M | 110.97M | 1402 | 213 |
| LPCDNet(MobileNetV2) | 1.82M | 764.23M | 962 | 131 |
| LPCDNet(ShuffleNetV2) | 783.25K | 328.23M | 2726 | 326 |
| **LPCDNet(WHU)** | **188.13K** | **118.93M** | **9163** | **1122** |
| **LPCDNet(GZLandslide)** | **266.57K** | **107.58M** | **8955** | **1655** |

\* FPS tested on Windows server.
\*\* FPS tested on NVIDIA Jetson AGX Orin.

From the results, we can observe that our LPCDNet has fewer parameters in comparison with the other patch-level CD methods and the MACs of it is only larger than RCVA, which is a conventional method without using neural networks. It should be noted that due to the different sensitivity values obtained on different datasets, the structure of LPCDNet on WHU-128 and GZLandslide-128 are not the same.

Concretely, on WHU-128 dataset, the channel number for four stages of LW-ResNet18 is 8, 36, 36, 33. While for GZLandslide-128, the channel number is 15, 8, 72 and 34.

As for FPS, LPCDNet achieves considerably higher value than other methods, with 9163/1122 on Windows server/AGX Orin on WHU-128 dataset and 8955/1655 on GZLandslide-128. On the whole, the better performance on three indicators compared to state-of-the-art methods fully reflects the great model efficiency of the proposed LPCDNet. Synthesizing the results of model efficiency and detection accuracy, we can conclude that our proposed LPCDNet achieves an excellent detection performance for patch-level CD with very low model complexity (<300K parameters and <120M MACs).

## C. Robustness to Registration Errors

As mentioned earlier, the use of pooling operation is able to reduce the impact of registration errors. To verify the robustness of the proposed network LPCDNet against registration errors, the detection performance of the network is tested under different levels of registration errors. Under given value of registration error, the process of generating unregistered image patch pairs is demonstrated in Fig. 6. When the registration error is equal to E pixels, for the input pair of image patches $A, B \epsilon \mathbb{R}^{H \times W}$, we capture the sub-region in the upper left and lower right corner of A and B respectively. Then two intermediate image patches $A', B' \epsilon \mathbb{R}^{(H-E) \times (W-E)}$ are obtained. Ultimately, the two patches are resized to the original size and a pair of unregistered image patches $A'', B'' \epsilon \mathbb{R}^{H \times W}$ are generated.

In experiment, the performance of our LPCDNet is measured under four different values of registration errors (i.e., 10, 20, 30, 40 pixels). The corresponding results are shown in Table III. As the registration error increases, PatchAcc, which measure the performance of patch-level CD, shows an obvious downward trend. Specifically, when the registration error is 10/20/30/40, the relative error of PatchAcc is 2.0%/4.5%/6.7%/8.7%. For input image patch of 128 * 128 size, the value of 40 pixels corresponds to a significant level of registration error, while the loss of detection accuracy is less than 10%. This fully demonstrates the robustness of the proposed network to registration errors. Meanwhile, for the detection rate of changed image patches which is quantified by the recall of positive class, the loss caused by the registration error is very limited. When the registration error is equal to 40, the value of Rec.(pos) only declines by 2.2%. In other words, although the registration error exists, the proposed LPCDNet is still able to select out most of the patches containing changed areas, which can ensure the overall detection accuracy of the whole CD process.

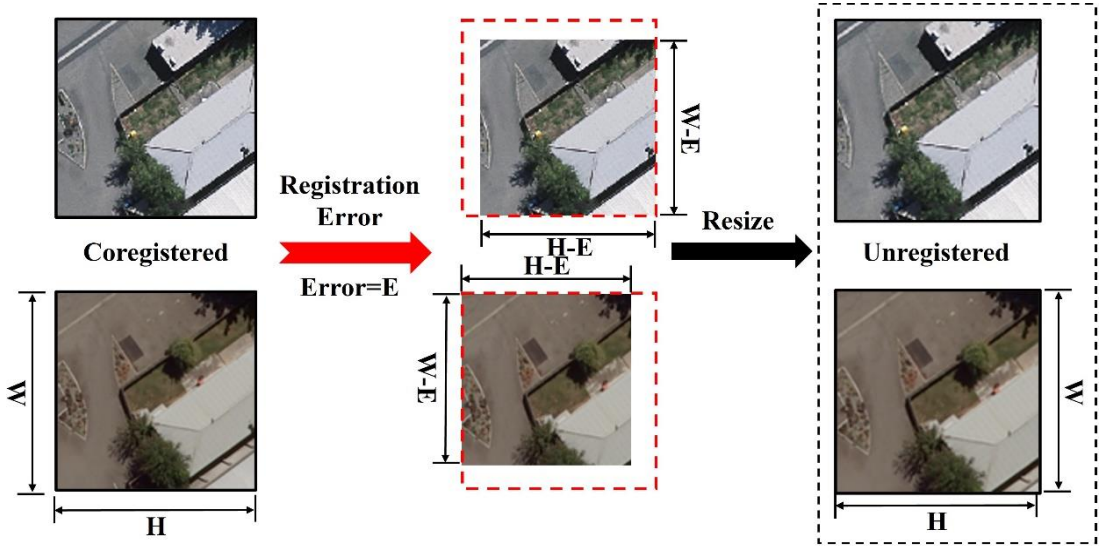

**Fig. 6. Process of generating unregistered image patch pairs based on given registration error.**

TABLE III
COMPARISON RESULTS UNDER DIFFERENT VALUES OF REGISTRATION ERROR ON WHU-128 DATASET.

| Registration Error | Rec.(neg) | Rec.(pos) | PatchAcc | Relative Error/% |
|---|---|---|---|---|
| 0 | 0.779 | 0.905 | 0.884 | 0 |
| 10 | 0.676 | 0.909 | 0.866 | 2.0 |
| 20 | 0.589 | 0.910 | 0.844 | 4.5 |
| 30 | 0.551 | 0.899 | 0.825 | 6.7 |
| 40 | 0.534 | 0.883 | 0.808 | 8.7 |

D. Ablation Studies

*1) Ablation on MLFC module:* To verify the strengths of the proposed MLFC module, we perform ablation study on the WHU-128 dataset, comparing the performance of network under different positions of feature compression. The ablation experiments are conducted before and after the application of network pruning method. The results are shown in Table IV and Table V.

In the experiment, two additional indicators, i.e., Kullback-Leibler Divergence (KLD) [81] and Jensen–Shannon Divergence (JSD) [82], are additionally introduced to describe the performance of the patch-level CD network. The calculation formulas for the two divergences are demonstrated below.

$$D_{KL}(p||q) = \sum_{i=1}^{N} p(x_i) \log\left(\frac{p(x_i)}{q(x_i)}\right)$$
$$D_{JS}(p||q) = \frac{1}{2}D_{KL}(p||\frac{p+q}{2}) + \frac{1}{2}D_{KL}(q||\frac{p+q}{2})$$
(6)

where $D_{KL}$ and $D_{JS}$ represent the KLD and JSD respectively. $\{p: p(x_i), i = 1, \dots, N\}$ and $\{q: q(x_i), i = 1, \dots, N\}$ denote the N-dimensional discrete probability distributions for positive class and negative class obtained by patch-level CD network. $\frac{p+q}{2}$ is the mean distribution of p and q. Both KLD and JSD are used to measure the difference between two distributions, where JSD is a variant of KLD to solve the problem of asymmetry. These two indicators are both non-negative, with the value of JSD ranging from 0 to 1. For JSD and KLD, larger value corresponds to greater difference in probability distribution between two classes.

TABLE IV

ABALTION STUDY OF DIFFERENT POSITIONS OF FEATURE COMPRESSION ON WHU-128 DATASET BEFORE APPLYING NETWORK PRUNING. THE TICK DENOTES USING POOLING-BASED FEATURE COMPRESSION AT CURRENT STAGE AND THE CROSS DENOTES THE ABSENCE OF COMPRESSION. THE VALUE OF PATCGACC AND F-SCORE ARE DESCRIBED IN PERCENTAGE(%). 1M IS EQUAL TO $10^6$ AND 1K IS EQUAL TO $10^3$.

| Stage0 | Stage1 | Stage2 | Stage3 | PatchAcc | F-score | KLD | JSD | Params.(M) | MACs(M) |
|---|---|---|---|---|---|---|---|---|---|
| √ | × | × | × | 86.29 | 68.16 | 1.78 | 0.32 | 4.89M | 940.36 |
| × | √ | × | × | 86.58 | 75.34 | 1.82 | 0.34 | 2.92M | 923.98 |
| × | × | √ | × | 86.76 | 74.58 | 1.66 | 0.33 | 2.80M | 921.61 |
| × | × | × | √ | 87.21 | 72.96 | 1.50 | 0.32 | 2.79M | 920.31 |
| √ | × | × | √ | 87.46 | 72.41 | 1.85 | 0.34 | 4.96M | 941.54 |
| √ | √ | √ | √ | **88.28** | **75.80** | **2.31** | **0.38** | 6.50M | 950.95 |

Table IV demonstrates the results on WHU-128 dataset under different positions of feature compression before applying network pruning strategy. Accordingly, the backbone network for feature extraction is original ResNet18 network and the channel number of four stages is 64, 64, 128 and 256 respectively. As shown in Table III, the detection performance after applying multi-layer feature compression is obviously superior to that of single-scale compression structure at each stage. To be specific, compared to the single-scale structure at stage 0/1/2/3, the introduction of MLFC module increases F-score by 7.64%/0.46%/1.22%/2.84% and PatchAcc by 1.99%/1.70%/1.52%/1.07% respectively. In addition, the KLD and JSD also improves, indicating the great discrimination ability of the global feature vector acquired by our proposed MLFC module. In view of the results in last two rows, it can be found that considering feature maps of all the stages to generate feature vectors yields significantly better detection results than only applying feature compression at the first and last stage. It suggests that the feature maps output by the middle layers of network also contain some important feature information that are conducive to identifying changed areas.

As for the model efficiency, even though the use of MLFC module leads to the increase of parameters, the absolute number of parameters is still very small. Meanwhile, the computational complexity of the algorithm, in terms of MACs, only exhibits slight increase. Therefore, the inference speed as well as inference time of the algorithm will not be affected.

TABLE V

ABALTION STUDY OF DIFFERENT POSITIONS OF FEATURE COMPRESSION ON WHU-128 DATASET AFTER NETWORK PRUNING. THE TICK DENOTES USING POOLING-BASED FEATURE COMPRESSION AT CURRENT STAGE AND THE CROSS DENOTES THE ABSENCE OF COMPRESSION.

**1M IS EQUAL TO $10^6$ AND 1K IS EQUAL TO $10^3$.**

| Stage0 | Stage1 | Stage2 | Stage3 | PatchAcc | F-score | KLD | JSD | Params.(K) | MACs(M) |
|---|---|---|---|---|---|---|---|---|---|
| √ | × | × | × | 85.87 | 68.82 | 2.00 | 0.32 | 162.54 | 118.49 |
| × | √ | × | × | 86.39 | 73.07 | 1.74 | 0.33 | 131.64 | 118.45 |
| × | × | √ | × | 85.69 | 71.66 | 1.49 | 0.30 | 129.65 | 118.22 |
| × | × | × | √ | 84.78 | 66.69 | 2.32 | 0.30 | 129.50 | 118.16 |
| √ | × | × | √ | 85.79 | 69.73 | 1.53 | 0.30 | 163.71 | 118.51 |
| √ | √ | × | × | 87.33 | 73.84 | 2.02 | 0.35 | 181.22 | 118.82 |
| × | √ | √ | × | 86.29 | 72.01 | 1.68 | 0.32 | 132.97 | 118.52 |
| × | × | √ | √ | 86.87 | 74.02 | 1.91 | 0.34 | 129.87 | 118.24 |
| √ | √ | √ | √ | **88.41** | **75.84** | **2.48** | **0.39** | 188.13 | 118.93 |

In Table V, we demonstrate the results of ablation experiment on WHU-128 dataset after network pruning. The backbone network used in experiment is LW-ResNet18. From Table V, we can draw conclusions similar to those in Table IV. After conducting network pruning, the detection accuracy of single-layer feature compression structure correspondingly decreases. Specifically, for single-scale structure at stage 0/1/2/3, the PatchAcc declines by 0.42%/0.19%/1.07%/2.43% after pruning. If the MLFC module is integrated, the detection performance of network will not be affected by pruning operation at all. Instead, the PatchAcc and F-score even rise by 0.13% and 0.04%, respectively. This fully reflects the effectiveness of the proposed multi-layer feature compression structure, as well as the robustness to network pruning. Under this circumstance, it is reasonable to conduct network pruning on the base network with MLFC module included.

To verify the effect of the proposed MLFC module, we also visualize the probability distributions obtained by the structure of single-layer and multi-layer feature compression. The visualization comparison on WHU-128 dataset is demonstrated in Fig. 7. The horizontal coordinate represents the probability of containing changed areas for image patch. From the figure, we can observe that the difference between the predicted probability distributions of the two classes is more pronounced for multi-scale feature compression structure compared to the single-scale structure, which helps to improve the final detection accuracy. Moreover, the feature information extracted by multi-scale structure is correspondingly more distinguished.

As shown in (a) and (b), for single-scale structure, the probability distribution obtained from low-level features is significantly more discriminative than that calculated from high-level features, resulting in higher detection accuracy. In comparison to the visualization result in (b), the range of values of predicted probability in (a) is smaller, and the tail probability of the distribution is considerably larger. This results in serious overlap between the probability distributions of two classes and greatly affects the accuracy of patch-level CD. For (b), (c) and (d), the visualization results of the predicted probability distributions appear to be similar. As the number of pooling layers increases, the trailing effect of the distribution gradually weakens, and the error for threshold-based decision decreases accordingly. This phenomenon accounts for the superior performance of multi-level feature compression based on multi-scale pooling.

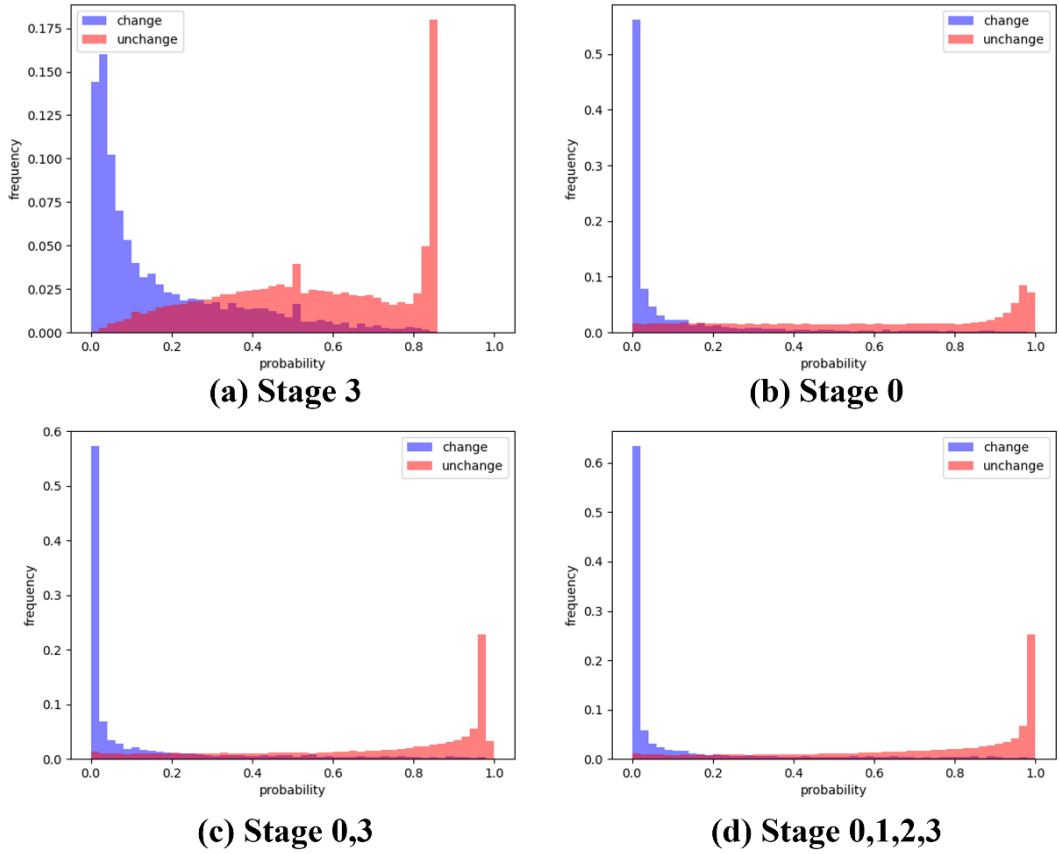

Fig. 7. Distribution of predicted probability for positive and negative class on WHU-128 test set under different positions of feature compression. The red color refers to the negative class(image patch with no changed areas) and the blue color refers to the positive class(image patch containing changed areas).

*2) Ablation on the sensitivity-guided network pruning method:* Table VI compares the detection accuracy and network efficiency of the pruned network under two strategies: sensitivity-guided network pruning and simple network pruning based on direct use of initial global pruning ratio. To fairly compare the two pruning strategies. To verify the effectiveness of the proposed sensitivity-guided pruning strategy, we compared two pruning strategies under similar value of MACs. For the simple pruning strategy where sensitivity is not used to correct the initial pruning rate, an appropriate global pruning ratio is chosen so that the MACs is larger than the sensitivity-guided strategy and as close to it as possible in the meanwhile. Under almost the same value of MACs, compared to the strategy which uses the same pruning ratio at each level of the network, the sensitivity-guided strategy yields a more efficient network structure and simultaneously improves detection accuracy. After modifying the initial pruning ratio using sensitivity, the PatchAcc improves by 0.70% and 0.81% on WHU-128 and GZLandslide-128 respectively. As for the F-score, it increases by 1.36% and 1.00%. In the meantime, the number of parameters after applying sensitivity-guided strategy is significantly smaller.

TABLE VI
THE RESULTS OF ABLATION EXPERIMENT FOR SENSITIVITY-GUIDED NETWORK PRUNING STRATEGY ON TWO DATASETS.

| Dataset | Initial Pruning Ratio | Sensitivity -guided | PatchAcc | F-score | Params. (K) | MACs (M) |
|---|---|---|---|---|---|---|
| WHU-128 | 0.340 | × | 87.71 | 74.48 | 382.36 | 125.92 |
|  | 0.125 | √ | **88.41** | **75.84** | **188.13** | **118.93** |
| GZLandslide-128 | 0.323 | × | 96.07 | 92.94 | 353.44 | 115.90 |
|  | 0.125 | √ | **96.88** | **93.94** | **266.57** | **107.58** |

*3) Ablation on the patch-level CD in the whole CD process:* In Section I, we have mentioned that the framework of combining patch-level CD and pixel-level CD is able to achieve more efficient inference with accurate detection results. To verify this statement, we conduct ablation experiment to validate the effectiveness of patch-level CD in improving efficiency of the complete CD process for large-scale remote sensing image. In the experiment, four pixel-level CD methods proposed recently are selected, including BIT [83], SNUNet [20], A2Net [31] and USSFC-Net [30]. Among them, SNUNet has high computational complexity, while A2Net and USSFC-Net are both lightweight methods. Using the original large image in [75], we measure the total processing time and final detection accuracy under the existence and absence of the proposed LPCDNet. As shown in Fig. 1, the whole process of CD of large-scale remote sensing image usually consists of three steps: image cropping, pixel-level CD and image merging. Because the time of cropping as well as merging is completely identical for four methods, we only consider pixel-level and patch-level CD process for test. The results of ablation study are demonstrated in Table VII. Note that the detection performance of the whole CD process is measured by overall accuracy (OA) and F1-score, which is different from the indicators for patch-level CD mentioned above.

From Table VII, it is observed that the inference time of the entire CD process is significantly shortened after introducing patch-level CD for pre-selection. To be specific, the processing time can be reduced to approximately 1/3 of the original value. In addition, with the application of patch-level CD, the ultimate pixel-level detection accuracy for the large image does not exhibit a significant decrease. For A2Net and USSFC-Net, F1-score only decreases by 0.5% and 0.4%. As for BIT and SNUNet, there is even a significant improvement of F1-score, increasing 1.9% and 1.4% respectively.

TABLE VII
ABALTION RESULTS OF THE PATCH-LEVEL CD IN THE COMPLETE PROCESS OF LARGE-SCALE REMOTE SENSING IMAGE CHANGE DETECTION. THE RESULTS ARE TESTED ON ORIGINAL LARGE IMAGE OF WHU-128 DATASET UNDER FOUR DIFFERENT PIXEL-LEVEL MODELS.

| Method | Recall | Precision | OA | F1-score | Time(s) |
|---|---|---|---|---|---|
| BIT [83] | 0.832 | 0.720 | 0.982 | 0.772 | 360.14 |
| Patch-level CD+BIT[83] | 0.825 | 0.759 | 0.984 | 0.791 | 162.72 |
| SNUNet [20] | 0.844 | 0.712 | 0.982 | 0.773 | 1375.96 |
| Patch-level CD+SNUNet [20] | 0.840 | 0.740 | 0.983 | 0.787 | 528.80 |
| A2Net [31] | 0.906 | 0.954 | 0.995 | 0.929 | 209.73 |
| Patch-level CD+A2Net [31] | 0.892 | 0.957 | 0.995 | 0.924 | 114.40 |
| USSFC-Net [30] | 0.875 | 0.932 | 0.993 | 0.903 | 524.97 |
| Patch-level CD+USSFC-Net [30] | 0.865 | 0.936 | 0.993 | 0.899 | 217.11 |

Compared with A2Net and USSFC-Net, there are considerably more false alarms in the detection results of BIT and SNUNet. Thus, if the pixel-level CD method is directly conducted on all of the image patches, some of the patches that do not contain changed areas will be wrongly detected to contain false alarm areas. This will have impact on the detection accuracy of the whole process. After incorporating patch-level CD, many of the unchanged image patches can be removed in advance, thereby reducing the occurrence of false alarms and improving the overall detection performance. Additionally, it should be noted that the values of F1-score for BIT and SNUNet are only 77.2% and 77.3% respectively, which are much lower than the original values measured in [83] and [20]. It is due to the difference between our WHU-128 dataset and the WHU-CD dataset used in [14] and [4], rather than errors in experiment.

## V. CONCLUSION

In this paper, we propose a lightweight patch-level CD network LPCDNet and use it to select the changed image patches cropped from the input large-scale optical remote sensing image. With this approach, the total processing time for CD of large image can be significantly shortened and the required computation and memory resources also greatly reduces, without considerable loss of detection accuracy. This provides the possibility for on-board processing of large-scale remote sensing images In LPCDNet, through the proposed sensitivity-guided channel pruning method, the lightweight backbone network LW-ResNet18 for feature extraction is built on basis of the ResNet18 network. In order to compress and fuse the feature maps of multiple levels, the MLFC module is designed and utilized to generate the feature vector with representations of multi-scale feature information for discrimination, in which the max-pooling operation is conducted on each stage of the backbone network.

Comparative experiments on two datasets for patch-level CD of optical remote sensing images (i.e., WHU-128 and GZLandslide-128) demonstrate that our LPCDNet achieves better detection performance than several recently advanced methods. For model efficiency, with extremely few parameters and MACs, the inference speed of LPCDNet is very fast, which is in accord with the requirements of patch-level CD task. Furthermore, the effectiveness of the MLFC module and the sensitivity-guided network pruning method are verified through the ablation study. Moreover, the comparison results also validate the effect of the proposed framework for large-scale image CD combining patch-level and pixel-level CD, compared to the concurrent mainstream framework with pure pixel-level CD.

For future research, considering that the proposed LPCDNet is targeted on optical remote sensing images, a natural idea is to focus on the patch-level CD of heterogeneous remote sensing images. In addition, enhancing the generalization performance of patch-level CD methods between different datasets and improving their detection accuracy in complex scenarios (e.g., cloud and fog occlusion, large registration errors) are also issues that need to be addressed.

# REFERENCES

[1] W. Shi, M. Zhang, H. Ke, X. Fang, Z. Zhan and S. Chen, "Landslide Recognition by Deep Convolutional Neural Network and Change Detection," in IEEE Transactions on Geoscience and Remote Sensing, vol. 59, no. 6, pp. 4654-4672, June 2021.

[2] Z. Wang, X. Wang, W. Wu and G. Li, "Continuous Change Detection of Flood Extents With Multisource Heterogeneous Satellite Image Time Series," in IEEE Transactions on Geoscience and Remote Sensing, vol. 61, pp. 1-18, 2023.

[3] X. Jiang, G. Li, X. -P. Zhang and Y. He, "A Semisupervised Siamese Network for Efficient Change Detection in Heterogeneous Remote Sensing Images," in IEEE Transactions on Geoscience and Remote Sensing, vol. 60, pp. 1-18, 2022.

[4] B. Demir, F. Bovolo and L. Bruzzone, "Updating Land-Cover Maps by Classification of Image Time Series: A Novel Change-Detection-Driven Transfer Learning Approach," in IEEE Transactions on Geoscience and Remote Sensing, vol. 51, no. 1, pp. 300-312, Jan. 2013.

[5] Z. Lei, T. Fang, H. Huo and D. Li, "Bi-Temporal Texton Forest for Land Cover Transition Detection on Remotely Sensed Imagery," in IEEE Transactions on Geoscience and Remote Sensing, vol. 52, no. 2, pp. 1227-1237, Feb. 2014.

[6] Y. Hu, Y. Dong, and Batunacun, "An automatic approach for land-change detection and land updates based on integrated NDVI timing analysis and the CVAPS method with GEE support," ISPRS J. Photogramm. Remote Sens., vol. 146, pp. 347–359, Dec. 2018.

[7] S. Ji, S. Wei and M. Lu, "Fully Convolutional Networks for Multisource Building Extraction From an Open Aerial and Satellite Imagery Data Set," in IEEE Transactions on Geoscience and Remote Sensing, vol. 57, no. 1, pp. 574-586, Jan. 2019.

[8] X. Huang, L. Zhang and T. Zhu, "Building Change Detection From Multitemporal High-Resolution Remotely Sensed Images Based on a Morphological Building Index," in IEEE Journal of Selected Topics in Applied Earth Observations and Remote Sensing, vol. 7, no. 1, pp. 105-115, Jan. 2014.

[9] C. Marin, F. Bovolo and L. Bruzzone, "Building Change Detection in Multitemporal Very High Resolution SAR Images," in IEEE Transactions on Geoscience and Remote Sensing, vol. 53, no. 5, pp. 2664-2682, May 2015.

[10] R. E. Kennedy et al., "Remote sensing change detection tools for natural resource managers: Understanding concepts and tradeoffs in the design of landscape monitoring projects," Remote Sens. Environ., vol. 113, no. 7, pp. 1382–1396, Jul. 2009.

[11] J. Jiang, J. Xiang, E. Yan, Y. Song and D. Mo, "Forest-CD: Forest Change Detection Network Based on VHR Images," in IEEE Geoscience and Remote Sensing

Letters, vol. 19, pp. 1-5, 2022.
[12] N. Zhang, X. Wei, H. Chen, and W. Liu, "FPGA implementation for CNN-based optical remote sensing object detection," Electronics, vol. 10, no. 3, p. 282, Jan. 2021.
[13] H. Mahendra, S. Mallikarjunaswamy, G. Siddesh, M. Komala, and N. Sharmila, "Evolution of real-time onboard processing and classification of remotely sensed data," Indian J. Sci. Technol., vol. 13, no. 20, pp. 2010–2020, May 2020.
[14] C. Vineyard, W. Severa, M. Kagie, A. Scholand, and P. Hays, "A resurgence in neuromorphic architectures enabling remote sensing computation," in Proc. IEEE Space Comput. Conf. (SCC), Jul. 2019, pp. 33–40.
[15] S. Liu and W. Luk, "Towards an efficient accelerator for DNN-based remote sensing image segmentation on FPGAs," in Proc. 29th Int. Conf. Field Program. Log. Appl. (FPL), Sep. 2019, pp. 187–193.
[16] C. Gonzalez, S. Bernabe, D. Mozos, and A. Plaza, "FPGA implementation of an algorithm for automatically detecting targets in remotely sensed hyperspectral images," IEEE J. Sel. Topics Appl. Earth Observ. Remote Sens., vol. 9, no. 9, pp. 4334–4343, Sep. 2016.
[17] N. Zhang, G. Wang, J. Wang, H. Chen, W. Liu and L. Chen, "All Adder Neural Networks for On-Board Remote Sensing Scene Classification," in IEEE Transactions on Geoscience and Remote Sensing, vol. 61, pp. 1-16, 2023, Art no. 5607916.
[18] R. C. Daudt, B. Le Saux, and A. Boulch, "Fully convolutional Siamese networks for change detection," in 2018 25th IEEE International Conference on Image Processing (ICIP). IEEE, 2018, pp. 4063–4067.
[19] D. Peng, Y. Zhang, and H. Guan, "End-to-end change detection for high resolution satellite images using improved unet++," Remote Sensing, vol. 11, no. 11, p. 1382, 2019.
[20] S. Fang, K. Li, J. Shao, and Z. Li, "Snunet-cd: A densely connected Siamese network for change detection of vhr images," IEEE Geoscience and Remote Sensing Letters, vol. 19, pp. 1–5, 2021.
[21] H. Jiang, X. Hu, K. Li, J. Zhang, J. Gong, and M. Zhang, "Pga-siamnet: Pyramid feature-based attention-guided Siamese network for remote sensing orthoimagery building change detection," Remote Sensing, vol. 12, no. 3, p. 484, 2020.
[22] H. Chen and Z. Shi, "A spatial-temporal attention-based method and a new dataset for remote sensing image change detection," Remote Sensing, vol. 12, no. 10, p. 1662, 2020.
[23] J. Chen, Z. Yuan, J. Peng, L. Chen, H. Huang, J. Zhu, Y. Liu, and H. Li,"Dasnet: Dual attentive fully convolutional Siamese networks for change detection in high-resolution satellite images," IEEE Journal of Selected Topics in Applied Earth Observations and Remote Sensing, vol. 14, pp. 1194–1206, 2020.
[24] J. Richards, "Thematic mapping from multitemporal image data using the principal components transformation," Remote Sensing of Environment, vol. 16, no. 1, pp. 35–46, 1984.
[25] S. Jin and S. A. Sader, "Comparison of time series tasseled cap wetness and the normalized difference moisture index in detecting forest disturbances," Remote Sensing of Environment, vol. 94, no. 3, pp. 364–372, 2005.

[26] K. Song, F. Cui, and J. Jiang, "An efficient lightweight neural network for remote sensing image change detection," Remote Sens., vol. 13, no. 24, p. 5152, Dec. 2021, doi: 10.3390/rs13245152.

[27] Dai, J.; Qi, H.; Xiong, Y.; Li, Y.; Zhang, G.; Hu, H.; Wei, Y. Deformable Convolutional Networks. In Proceedings of the 2017 IEEE International Conference on Computer Vision (ICCV), Venice, Italy, 22–29 October 2017; pp. 764–773.

[28] B. Liu, H. Chen, Z. Wang, W. Xie and L. Shuai, "LSNET: Extremely Light-Weight Siamese Network for Change Detection of Remote Sensing Image," IGARSS 2022 - 2022 IEEE International Geoscience and Remote Sensing Symposium, Kuala Lumpur, Malaysia, 2022, pp. 2358-236.

[29] A. Codegoni, G. Lombardi, and A. Ferrari, "Tinycd: a (not so) deep learning model for change detection," Neural Computing and Applications, vol. 35, no. 11, pp. 8471–8486, 2023.

[30] T. Lei et al., "Ultralightweight Spatial–Spectral Feature Cooperation Network for Change Detection in Remote Sensing Images," in IEEE Transactions on Geoscience and Remote Sensing, vol. 61, pp. 1-14, 2023, Art no. 4402114.

[31] Z. Li et al., "Lightweight Remote Sensing Change Detection With Progressive Feature Aggregation and Supervised Attention," in IEEE Transactions on Geoscience and Remote Sensing, vol. 61, pp. 1-12, 2023, Art no. 5602812.

[32] M. Sandler, A. Howard, M. Zhu, A. Zhmoginov, and L.-C. Chen,"Mobilenetv2: Inverted residuals and linear bottlenecks," in Proceedings of the IEEE conference on computer vision and pattern recognition, 2018, pp. 4510–4520.

[33] K. He, X. Zhang, S. Ren, and J. Sun, "Deep residual learning for image recognition," in Proceedings of the IEEE conference on computer vision and pattern recognition, 2016, pp. 770–778.

[34] K. Simonyan and A. Zisserman, "Very deep convolutional networks for large-scale image recognition," arXiv preprint arXiv:1409.1556, 2014.

[35] O. Ronneberger, P. Fischer, and T. Brox, "U-net: Convolutional networks for biomedical image segmentation," in Proc. Int. Conf. Med. Image Comput. Comput.-Assisted Intervention, 2015, pp. 234–241.

[36] T. Bao, C. Fu, T. Fang, and H. Huo, "Ppcnet: A combined patch-level and pixel-level end-to-end deep network for high-resolution remote sensing image change detection," IEEE Geoscience and Remote Sensing Letters, vol. 17, no. 10, pp. 1797–1801, 2020.

[37] F. Rahman, B. Vasu, J. Van Cor, J. Kerekes, and A. Savakis, "Siamese network with multi-level features for patch-based change detection in satellite imagery," in 2018 IEEE Global Conference on Signal and Information Processing (GlobalSIP). IEEE, 2018, pp. 958–962.

[38] S. Saha and X. X. Zhu, "Patch-level unsupervised planetary change detection," IEEE Geoscience and Remote Sensing Letters, vol. 19, pp. 1–5, 2021.

[39] S. Han, X. Liu, H. Mao, J. Pu, A. Pedram, M. A. Horowitz, and W. J. Dally, "Eie: Efficient inference engine on compressed deep neural network," ACM SIGARCH Computer Architecture News, vol. 44, no. 3, pp. 243–254, 2016.

[40] S. Han, J. Pool, J. Tran, and W. J. Dally, "Learning both weights and connections

for efficient neural networks," in Proc. 28th Int. Conf. Neural Inf. Process. Syst., Montreal, QC, Canada, Dec. 2015, pp. 1135–1143.
[41] S. Han, H. Mao, and W. J. Dally. "Deep Compression: Compressing Deep Neural Networks With Pruning, Trained Quantization and Huffman Coding." Oct. 2015. [Online]. Available: https://arxiv.org/abs/1510.00149.
[42] Z. Liu, M. Sun, T. Zhou, G. Huang, and T. Darrell, "Rethinking the value of network pruning," arXiv preprint arXiv:1810.05270, 2018.
[43] Hao Li, Asim Kadav, Igor Durdanovic, Hanan Samet, and Hans Peter Graf. Pruning filters for efficient convnets. In International Conference of Learning Representation (ICLR), 2017.
[44] Yang He, Guoliang Kang, Xuanyi Dong, Yanwei Fu, and Yi Yang. Soft filter pruning for accelerating deep convolutional neural networks. In Proceedings of the Twenty-Seventh International Joint Conference on Artificial Intelligence, IJCAI-18, pages 2234–2240, 2018.
[45] Yang He, Ping Liu, Ziwei Wang, Zhilan Hu, and Yi Yang. Filter pruning via geometric median for deep convolutional neural networks acceleration. In Computer Vision and Pattern Recognition (CVPR), 2019.
[46] Y. Lu, M. Gong, Z. Hu, W. Zhao, Z. Guan and M. Zhang, "Energy-Based CNN Pruning for Remote Sensing Scene Classification," in IEEE Transactions on Geoscience and Remote Sensing, vol. 61, pp. 1-14, 2023.
[47] Mingbao Lin, Rongrong Ji, Yan Wang, Yichen Zhang, Baochang Zhang, Yonghong Tian, and Ling Shao. Hrank: Filter pruning using high-rank feature map. In Proceedings of the IEEE/CVF Conference on Computer Vision and Pattern Recognition, pages 1529–1538, 2020.
[48] Y. Sui, M. Yin, Y. Xie, H. Phan, S. Aliari Zonouz, and B. Yuan, "CHIP: CHannel independence-based pruning for compact neural networks," in Proc. NeurIPS, 2021, pp. 1–13.
[49] J.-H. Luo and J. Wu, "An entropy-based pruning method for CNN compression," 2017, arXiv:1706.05791.
[50] Z. Liu, Z. Shen, M. Savvides, and K.-T. Cheng, "ReActNet: Towards precise binary neural network with generalized activation functions," 2020. [Online]. Available: arxiv.abs/2003.03488.
[51] A. Polino, R. Pascanu, and D. Alistarh, "Model compression via distillation and quantization," 2018, arXiv:1802.05668.
[52] W. Chen, J. T. Wilson, S. Tyree, K. Q. Weinberger, and Y. Chen, "Compressing neural networks with the hashing trick," in Proc. Int. Conf. Mach. Learn., Lille, France, Jul. 2015.
[53] Y. Idelbayev and M. A. Carreira-Perpinan, "Low-rank compression of neural nets: Learning the rank of each layer," in Proc. IEEE Conf. Comput. Vis. Pattern Recognit., 2020, pp. 8049–8059.
[54] X. Zhang, J. Zou, X. Ming, K. He, and J. Sun, "Efficient and accurate approximations of nonlinear convolutional networks," in Proc. IEEE Conf. Comput. Vis. Pattern Recognit. (CVPR), Jun. 2015, pp. 1984–1992.
[55] Y. W. Li, S. H. Gu, C. Mayer, L. Van Gool, and R. Timofte, "Group sparsity: The

hinge between filter pruning and decomposition for network compression," in Proc. IEEE/CVF Conf. Computer Vision and Pattern Recognition, Seattle, USA, 2020, pp. 8015−8024.

[56] D. Walawalkar, Z. Shen, and M. Savvides, "Online ensemble model compression using knowledge distillation," in Proc. Eur. Conf. Comput. Vis., Springer, 2020, pp. 18–35.

[57] T. Li, J. Li, Z. Liu, and C. Zhang, "Few sample knowledge distillation for efficient network compression," in Proc. IEEE/CVF Conf. Comput. Vis. Pattern Recognit., 2020, pp. 14639–14647.

[58] J. -J. Liu, Q. Hou, Z. -A. Liu and M. -M. Cheng, "PoolNet+: Exploring the Potential of Pooling for Salient Object Detection," in IEEE Transactions on Pattern Analysis and Machine Intelligence, vol. 45, no. 1, pp. 887-904, 1 Jan. 2023.

[59] Yann LeCun, Bernhard E Boser, John S Denker, Donnie Henderson, Richard E Howard, Wayne E Hubbard, and Lawrence D Jackel. Handwritten digit recognition with a back-propagation network. In Advances in neural information processing systems, pages 396–404, 1990.

[60] Y. Lecun, L. Bottou, Y. Bengio and P. Haffner, "Gradient-based learning applied to document recognition," in Proceedings of the IEEE, vol. 86, no. 11, pp. 2278-2324, Nov. 1998.

[61] M. A. Ranzato, Y. L. Boureau, and Y. Lecun, "Sparse feature learning for deep belief networks," in Proc. Int. Conf. Neural Inf. Process. Syst., 2007, pp. 1185–1192.

[62] C.-Y. Lee, P. Gallagher, and Z. Tu, "Generalizing pooling functions in CNNs: Mixed, gated, and tree," IEEE Trans. Pattern Anal. Mach. Intell., vol. 40, no. 4, pp. 863–875, Apr. 2018.

[63] M. D. Zeiler and R. Fergus, "Stochastic pooling for regularization of deep convolutional neural networks," in Proc. ICLR, 2013.

[64] M. Lin, Q. Chen, and S. Yan, "Network in network," in Proc. Int. Conf. Learn. Representations, 2014. [Online]. Available: https://arxiv.org/abs/1312.4400.

[65] A. Stergiou, R. Poppe, and G. Kalliatakis, "Refining activation downsampling with SoftPool," 2021, arXiv:2101.00440.

[66] Z. Gao, L. Wang, and G. Wu, "LIP: Local importance-based pooling," in Proc. Int. Conf. Comput. Vis., 2019, pp. 3355–3364.

[67] Q. Hou, L. Zhang, M.-M. Cheng, and J. Feng, "Strip pooling: Rethinking spatial pooling for scene parsing," in Proc. IEEE Conf. Comput. Vis. Pattern Recognit., 2020, pp. 4003–4012.

[68] A. G. Howard, M. Zhu, B. Chen, D. Kalenichenko, W. Wang, T. Weyand, M. Andreetto, and H. Adam, "Mobilenets: Efficient convolutional neural networks for mobile vision applications," arXiv preprint arXiv:1704.04861, 2017.

[69] A. Howard, M. Sandler, G. Chu, L.-C. Chen, B. Chen, M. Tan, W. Wang, Y. Zhu, R. Pang, V. Vasudevan et al., "Searching for mobilenetv3," in Proceedings of the IEEE/CVF international conference on computer vision, 2019, pp. 1314–1324.

[70] K. Han, Y. Wang, Q. Tian, J. Guo, C. Xu, and C. Xu, "Ghostnet: More features from cheap operations," in Proceedings of the IEEE/CVF conference on computer vision and pattern recognition, 2020, pp. 1580–1589.

[71] N. Ma, X. Zhang, H.-T. Zheng, and J. Sun, "Shufflenet v2: Practical guidelines for efficient cnn architecture design," in Proceedings of the European conference on computer vision (ECCV), 2018, pp. 116–131.

[72] B. Wu, A. Wan, X. Yue, P. Jin, S. Zhao, N. Golmant, A. Gholaminejad, J. Gonzalez, and K. Keutzer, "Shift: A zero flop, zero parameter alternative to spatial convolutions," in Proceedings of the IEEE conference on computer vision and pattern recognition, 2018, pp. 9127–9135.

[73] X. Guo, B. Hou, B. Ren, Z. Ren and L. Jiao, "Network Pruning for Remote Sensing Images Classification Based on Interpretable CNNs," in IEEE Transactions on Geoscience and Remote Sensing, vol. 60, pp. 1-15, 2022.

[74] Y. He, X. Zhang, and J. Sun, "Channel pruning for accelerating very deep neural networks," in Proc. IEEE Int. Conf. Comput. Vis. (ICCV), Oct. 2017, pp. 1389–1397.

[75] S. Ji, S. Wei, and M. Lu, "Fully convolutional networks for multisource building extraction from an open aerial and satellite imagery data set," IEEE Transactions on Geoscience and Remote Sensing, vol. 57, no. 1, pp. 574–586, 2018.

[76] P. Xu et al., "On-board real-time ship detection in HISEA-1 SAR images based on CFAR and lightweight deep learning," Remote Sens., vol. 13, no. 10, 2021, Art. no. 1995.

[77] L. Bai, Q. Liu, C. Li, C. Zhu, Z. Ye, and M. Xi, "A lightweight and multiscale network for remote sensing image scene classification," IEEE Geoscience and Remote Sensing Letters, vol. 19, pp. 1–5, 2021.

[78] B. Zhang, Y. Zhang, and S. Wang, "A lightweight and discriminative model for remote sensing scene classification with multidilation pooling module," IEEE Journal of Selected Topics in Applied Earth Observations and Remote Sensing, vol. 12, no. 8, pp. 2636–2653, 2019.

[79] F. Thonfeld, H. Feilhauer, M. Braun, and G. Menz, "Robust change vector analysis (rcva) for multi-sensor very high resolution optical satellite data," International Journal of Applied Earth Observation and Geoinformation, vol. 50, pp. 131–140, 2016.

[80] M. Wu, G. Cheng, X. Yao, X. Qian, J. Han, and L. Guo, "Performance comparison of two pooling strategies for remote sensing image scene classification," in IGARSS 2019-2019 IEEE International Geoscience and Remote Sensing Symposium. IEEE, 2019, pp. 3037–3040.

[81] S. Kullback and R. A. Leibler, "On information and sufficiency," The annals of mathematical statistics, vol. 22, no. 1, pp. 79–86, 1951.

[82] J. Lin, "Divergence measures based on the shannon entropy," IEEE Transactions on Information theory, vol. 37, no. 1, pp. 145–151, 1991.

[83] H. Chen, Z. Qi, and Z. Shi, "Remote sensing image change detection with transformers," IEEE Transactions on Geoscience and Remote Sensing, vol. 60, pp. 1–14, 2022.